\documentclass{article} 
\usepackage[final]{colm2025_conference}

\usepackage{microtype}
\usepackage{hyperref}
\usepackage{url}
\usepackage{booktabs}

\usepackage{wrapfig}

\usepackage{colortbl}

\usepackage{graphicx}
\usepackage[utf8]{inputenc} 
\usepackage[T1]{fontenc}    
\usepackage{hyperref}       
\usepackage{url}            
\usepackage{booktabs}       
\usepackage{amsfonts}       
\usepackage{nicefrac}       
\usepackage{microtype}      
\usepackage{xcolor}         

\usepackage{multirow}

\usepackage{enumitem}
\usepackage{bm}%
\usepackage{amsmath}%

\usepackage{amssymb}%
\usepackage{amsthm}
\usepackage{colonequals}
\usepackage{psfrag}
\usepackage{mathrsfs}

\usepackage{booktabs,siunitx}

\usepackage{algorithm}
\usepackage{algpseudocode}

\usepackage{comment}

\usepackage{lineno}

\definecolor{darkblue}{rgb}{0, 0, 0.5}
\hypersetup{colorlinks=true, citecolor=darkblue, linkcolor=darkblue, urlcolor=darkblue}

\newtheorem{thm}{Theorem}
\newtheorem{lem}{Lemma}

\newtheorem{prop}{Proposition}

\newtheorem{cor}{Corollary}

\theoremstyle{definition}

\newtheorem{defn}{Definition}

\newtheorem{rem}{Remark}

\newcommand{\ba}{\bm{a}}
\newcommand{\bb}{\bm{b}}

\newcommand{\bq}{\bm{q}}

\newcommand{\bx}{\bm{x}}
\newcommand{\bd}{\bm{d}}

\newcommand{\bv}{\bm{v}}

\newcommand{\bw}{\bm{w}}
\newcommand{\bh}{\bm{h}}

\newcommand{\bk}{\bm{k}}

\newcommand{\bH}{\mathbf{H}}

\newcommand{\bC}{\mathbf{C}}
\newcommand{\bI}{\mathbf{I}}
\newcommand{\bM}{\mathbf{M}}
\newcommand{\bN}{\mathbf{N}}
\newcommand{\bW}{\mathbf{W}}
\newcommand{\bA}{\mathbf{A}}
\newcommand{\bB}{\mathbf{B}}
\newcommand{\bR}{\mathbf{R}}
\newcommand{\bP}{\mathbf{P}}
\newcommand{\bQ}{\mathbf{Q}}
\newcommand{\bD}{\mathbf{D}}

\newcommand{\bX}{\mathbf{X}}
\newcommand{\bV}{\mathbf{V}}
\newcommand{\bK}{\mathbf{K}}
\newcommand{\bS}{\mathbf{S}}
\newcommand{\bT}{\mathbf{T}}
\newcommand{\bO}{\mathbf{O}}

\newcommand{\blambda}{\pmb{\lambda}}

\newcommand{\bdelta}{\pmb{\delta}}

\newcommand{\bSigma}{\pmb{\Sigma}}

\newcommand{\diag}{\mathsf{diag}}

\newcommand{\Reals}{\mathbb{R}}
\newcommand{\defined}{\triangleq}

\newcommand{\sto}{\mbox{\normalfont s.t.}}

\title{SQuat: \underline{S}ubspace-orthogonal KV Cache \underline{Qua}n\underline{t}ization}

\author{Hao Wang$^{*}$, Ligong Han\thanks{Equal contribution. Correspondence to: \texttt{\{hao-wang,lihan\}@redhat.com}}, ~Kai Xu, Akash Srivastava\\
Red Hat AI Innovation 
}

\begin{document}

\ifcolmsubmission
\linenumbers
\fi

\maketitle

\begin{abstract}
The key-value (KV) cache accelerates LLMs decoding by storing KV tensors from previously generated tokens. It reduces redundant computation at the cost of increased memory usage. To mitigate this overhead, existing approaches compress KV tensors into lower-bit representations; however, quantization errors can accumulate as more tokens are generated, potentially resulting in undesired outputs. In this paper, we introduce \textbf{SQuat} (\textbf{S}ubspace-orthogonal KV cache \textbf{qua}n\textbf{t}ization). It first constructs a subspace spanned by query tensors to capture the most critical task-related information. During key tensor quantization, it enforces that the difference between the (de)quantized and original keys remains orthogonal to this subspace, minimizing the impact of quantization errors on the attention mechanism's outputs. SQuat requires no model fine-tuning, no additional calibration dataset for offline learning, and is grounded in a theoretical framework we develop. Through numerical experiments, we show that our method reduces peak memory by $\mathbf{2.17\times} \sim \mathbf{2.82\times}$, improves throughput by $\mathbf{2.45\times} \sim \mathbf{3.60\times}$, and achieves more favorable benchmark scores than existing KV cache quantization algorithms. Our code is available at: \href{https://github.com/Red-Hat-AI-Innovation-Team/SQuat}{https://github.com/Red-Hat-AI-Innovation-Team/SQuat}
\end{abstract}

\section{Introduction}

The future of AI technology should be accessible to everyone and operate efficiently. Just as computers evolved from bulky, expensive machines into widely available, affordable tools, AI is undergoing a similar transformation. Today, large language models (LLMs) are increasingly used in many applications, yet challenges remain. Unlike traditional software applications that execute tasks almost instantly, LLMs---even during inference---require significant computational resources to process user queries. To reduce compute costs in auto-regressive decoding, causal language models often use the key-value (KV) cache. This cache stores previously computed keys and values for the attention mechanism, allowing the model to avoid redundant calculations.  In essence, it is just like how humans remember the context of a conversation---we do not start from scratch with every reply, but build on what was said before.

A major bottleneck introduced by the KV cache is the increased GPU memory consumption and the latency from frequent data transfers between memory and compute units~\citep{lienhart2024llminf,verma2023master}. The challenge is amplified in modern reasoning models \citep{openaio1,guo2025deepseek}, which often require long responses to ``think'' through problems before arriving at a final answer. Moreover, inference-time scaling methods further exacerbate these issues by increasing context lengths or the number of concurrent decoding paths \citep{lightman2023lets,rushspeculations2024}. 
To address this challenge, a burgeoning line of research investigates strategies to optimize KV cache management, mitigate memory constraints, and accelerate inference speed \citep[see e.g.,][]{kwon2023efficient,dubey2024llama,liu2024deepseek}. Within this effort, KV cache quantization is a crucial approach that compresses stored keys and values into lower-precision formats. It eliminates the need for re-training or fine-tuning LLMs, preserves full-context information, and remains compatible with methods like token pruning and model compression.

Existing approaches to KV cache quantization typically treat it as a lossy data compression problem \citep[see e.g.,][]{yang2024lossless,liu2024kivi,kang2024gear,hooper2024kvquant}. They aim to minimize the discrepancy between full-precision KV tensors (e.g., FP16) and their low-bit representations (e.g., INT2) without explicitly accounting for their effect on the model’s next-token prediction. Consequently, while they preserve the numerical fidelity of KV tensors to some extent, task-critical information---necessary for accurately responding to user queries---may be lost. Moreover, quantization errors from early tokens can accumulate and amplify during decoding, causing later tokens to deviate from expected outputs and eventually leading to degraded response quality or misalignment with user queries. This motivates a fundamental question:
\begin{center}
    \emph{How can we preserve task-relevant information during KV cache quantization to ensure that the LLM generates outputs as if using full-precision KV tensors?}
\end{center}

\begin{figure*}[t]
    \centering
    \includegraphics[width=0.95\linewidth]{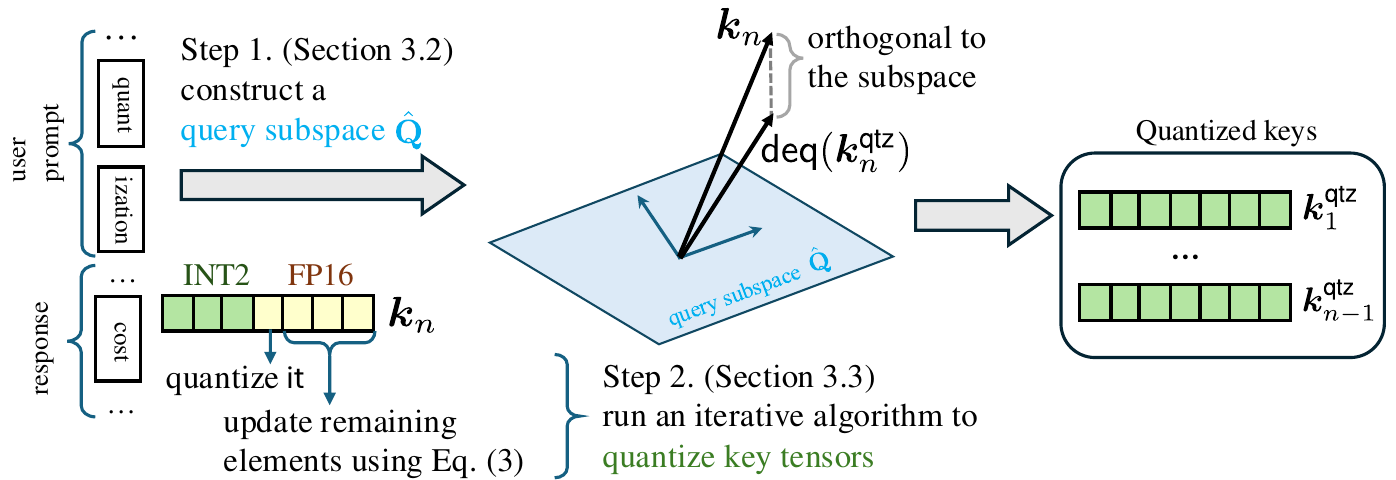}
    \caption{The pipeline of SQuat for quantizing key tensors. We first identify a subspace $\hat{\bQ}$ spanned by query tensors that carry the most relevant task information necessary for answering the user question. Then we quantize each key tensor $\bk_n$ element by element (or block by block) and update the remaining elements using Eq.~\eqref{eq::opt_update} at each iteration. This update ensures that the difference between the (de)quantized and original key remains as orthogonal to the query subspace as possible, preserving critical information for responding to the user question.
    }
    \label{fig:pipeline}
\end{figure*}

In this paper, we introduce \textbf{SQuat} (\textbf{S}ubspace-orthogonal KV cache \textbf{qua}n\textbf{t}ization), a new approach for KV cache quantization. 
It is motivated by our two observations: 
\begin{itemize}[leftmargin=1em]
\item \emph{Observation 1.} The attention scores in transformer architectures are calculated by taking the inner product between the query tensors of the new token and the key tensors of all past tokens. Thus, when quantizing key tensors, the primary objective should be to preserve their inner products with future tokens' query tensors, rather than merely minimizing the difference between the original and (de)quantized key tensors.

\item \emph{Observation 2.} Query tensors tend to lie within a subspace whose dimension is significantly smaller than the hidden dimension. Hence, even without knowing future tokens' query tensors at the time of quantizing the current token's key tensors, we can still leverage this subspace to guide the quantization process. 
\end{itemize}
Building on these insights, \emph{SQuat} first constructs a task-relevant subspace using the query tensors from all tokens in user-provided prompts. It then quantizes each token's key tensors, ensuring that the residuals (i.e., differences between the (de)quantized and original key tensors) stays as orthogonal to this subspace as possible, reducing the quantization error's effect on critical task information (see Figure~\ref{fig:pipeline} for our pipeline). \emph{SQuat} requires no model training, no additional calibration data for offline learning, and is grounded in a theoretical framework we develop.

\emph{SQuat} is based on an optimization that balances the trade-off between maintaining quantization fidelity and keeping residuals orthogonal to the query subspace. We introduce an iterative algorithm that approximately solves this optimization problem. At each step, an element (or block of elements) of the key tensors is quantized, followed by an update to the remaining elements. This update minimizes the quantization error’s impact on inner products with vectors in the query subspace. We prove that the optimal update rule has a closed-form expression and present an efficient algorithm for computing it. This effort significantly reduces computational complexity and enables \emph{SQuat} to run on-the-fly.

We conduct extensive numerical experiments to evaluate \emph{SQuat} and compare it against tuning-free baselines. Specifically, we apply these methods to quantize the KV cache of four LLMs: \texttt{Llama-2-7B} \citep{touvron2023llama}, \texttt{Llama-3.1-8B-Instruct} \citep{dubey2024llama}, \texttt{Mistral-7B-Instruct-v0.3} \citep{jiang2023mistral7b}, and \texttt{DeepSeek-R1-Distill-Llama-8B} \citep{guo2025deepseek}. We evaluate their performance across two benchmark families: (1) a set of challenging reasoning tasks that often need long responses before reaching the final answer, and (2) LongBench tasks that focus on long-context understanding, such as document QA and summarization, which require long input windows \citep{bai2023longbench}. Together, these benchmarks cover 14 tasks. \emph{SQuat} achieve more favorable performance than existing tuning-free baselines. When applied to the \texttt{Llama-2-7B} model, SQuat reduces peak memory usage by $\mathbf{2.17\times} \sim \mathbf{2.82\times}$ compared to the default FP16 format, resulting in a throughput improvement of $\mathbf{2.45\times} \sim \mathbf{3.60\times}$.

\subsection{Related Work}

\paragraph{KV cache management.} Recent research has explored advanced KV cache management techniques to optimize inference efficiency and reduce memory usage in LLMs. 
One strategy focuses on token pruning, eviction, or merging \citep{xiao2023efficient, beltagy2020longformer, kim2022learned, ge2023model, liu2024scissorhands, han2023lm, zhang2023h2o, lee2025infini, wang2024model,wan2024d2o,zhang2024cam,jiang2025minference}, where older or less relevant tokens are removed from the KV cache, or similar and redundant tokens are merged. Some methods \citep{cai2024pyramidkv,li2024snapkv} allow for selective eviction of KV tensors across different attention layers rather than uniformly pruning KV tensors for the same tokens at all layers. A related technique \citep{xu2024think} prunes specific key tensor channels instead of entire tokens, leveraging query information to guide the pruning process.
Beyond pruning-based methods, there are other techniques to reduce the memory overhead of storing the KV cache. For example, \citet{tang2024quest,hooper2024squeezed,ribar2023sparq} propose storing the full KV cache but dynamically load only the relevant keys and values during inference. These approaches, along with advancements in system and architecture design, are orthogonal to our method on KV cache quantization and can potentially be combined together to enhance GPU utilization.

\paragraph{KV cache compression.} 
There has been active research on KV cache quantization \citep{tao2024asymkv,yang2024lossless,byun2024hessian,zhang2024csr,liu2024kivi,kang2024gear,hooper2024kvquant,sheng2023flexgen,zandieh2024qjl,he2025zipcache,yue2024wkvquant,liu2024intactkv,chang2024palu,zhang2024zero,duanmu2024skvq,zhao2024atom}. For example, \cite{xiao2023smoothquant} \cite{sheng2023flexgen} introduce 8-bit and 4-bit group-wise quantization for both model weights and KV cache. 
More recently, \citet{liu2024kivi,kang2024gear,hooper2024kvquant} explore per-channel quantization for key tensors and per-token quantization for value tensors, as key tensors show significant variation across channels, whereas values do not. 
Similarly, \citet{duanmu2024skvq} propose rearranging KV cache channels to enhance similarity within quantization groups. 
\citet{hooper2024kvquant} further introduce pre-RoPE quantization, outlier isolation, and non-uniform quantization. \citet{kim2024lexico} develop an input-independent dictionary and represent KV tensors as sparse linear combinations of its elements. 
These advancements enable 2-bit quantization while maintaining performance comparable to non-quantized baselines. 
Among these approaches, \citet{liu2024kivi,kang2024gear,he2025zipcache}, like our method, do not require fine-tuning or a calibration dataset. \citet{liu2024kivi} present a hardware-friendly GPU implementation, inspiring a new quantization method integrated into Hugging Face Transformers \citep{turganbay2024unlock}. \citet{kang2024gear} reduce quantization error by storing a low-rank matrix and a sparse matrix. \citet{he2025zipcache} consider first identifying salient tokens and then applying mixed-precision quantization to compress the KV cache. As we will discuss in Section~\ref{sec::motivation}, these compression-based approaches do not explicitly account for quantization errors' impact on attention outputs. In contrast, our method preserves the inner product between key tensors and future tokens' query tensors, minimizing quantization-induced deviations in LLM outputs. A comprehensive comparison with these methods will be provided in our numerical experiments (Section~\ref{sec::exp}).

\paragraph{Efficient system and architecture designs.} Optimizing computer systems and transformer architectures for efficient LLM inference has been an active research area. One line of research focuses on system design for optimizing KV cache management on GPUs \citep{jin2023s,yu2022orca,FasterTransformer23}. Among them, \citet{kwon2023efficient} propose paged attention, a block-based dynamic memory allocation system that serves as the foundation of vLLM, a widely adopted library for LLM inference and serving. Another research direction explores modifications to attention mechanisms in transformer-based LLMs to reduce KV cache size \citep{child2019generating,katharopoulos2020transformers,choromanski2020rethinking,wu2024layer,liu2024minicache,liu2024deepseek,yuan2025natives,yu2024effectively}. For example, multi-query and grouped-query attention mechanisms maintain multiple query heads while sharing a single set of KV pairs across all heads or within groups of heads \citep{ainslie2023gqa,shazeer2019fast}. This architecture has been incorporated into many open-source LLMs, including Llama \citep{dubey2024llama} and PaLM \citep{chowdhery2023palm}.

\paragraph{Model compression.} The study of pruning, quantization, and sparsification in neural networks has a long history \citep{gupta2015deep,frankle2018lottery,frantar2022optimal}, originating from early efforts \citep{lecun1989optimal,hassibi1993optimal} to optimize computational efficiency and memory usage while preserving accuracy. Recently, these techniques have gained increasing attention in the context of LLMs, which have grown to hundreds of billions of parameters, demanding substantial compute resources for both training and inference \citep{frantar2023sparsegpt,sun2023simple,frantar2022gptq,xiao2023smoothquant,yao2022zeroquant,dettmers2022gpt3,kurtic2023ziplm,malinovskii2024pushing,ashkboos2025quarot}. 
Our work focuses on quantizing the KV cache rather than model weights. Unlike static weight matrices, which can be pre-quantized and stored before inference, KV tensors are generated dynamically as new tokens arrive. This streaming nature demands quantization to be performed on-the-fly, requiring methods that minimize computational overhead and latency to ensure efficient decoding.

\section{Background}

\paragraph{(Masked) Attention mechanism.} 
For the sake of illustration, we assume batch size is $1$ and the attention mechanism has a single head. 
A transformer consists of $L$ layers and each layer contains a multi-head attention and a feed-forward network. Suppose the input token embeddings are: $\bX \in \Reals^{n \times d}$. 
\begin{align*}
    \text{Query matrix: }\bQ = \bX \bW^Q,\quad
    \text{Key matrix: }\bK = \bX \bW^K,\quad
    &\text{Value matrix: }\bV = \bX \bW^V,
\end{align*}
where $\bW^Q, \bW^K, \bW^V \in \Reals^{d\times d}$ are the projection matrices. The attention mechanism outputs: $\mathsf{Attention}(\bQ, \bK, \bV) = \mathsf{softmax}\left(\frac{\bQ \bK^T}{\sqrt{d}} + \bM \right)\bV$, where the mask $\bM$ is a strictly upper triangular matrix, with zeros on and below the diagonal and $-\infty$ in every element above the diagonal.

\paragraph{Inference-time workflow.} In the \emph{prefill phase}, let $\bX_0 \in \Reals^{l_{\text{prompt}} \times d}$ represent the input tensor, where $l_{\text{prompt}}$ is the length of the user's input prompt and $d$ denotes the model's hidden size. During this phase, the key ($\bK$) and value ($\bV$) tensors are computed and cached. This is followed by the \emph{decoding phase}, in which the LLM predicts the next token sequentially, conditioned on the tokens generated so far. Note that LLMs only compute the query, key, and value for the newly generated token. The new key and value are then concatenated with the previously cached tensors. This caching mechanism significantly reduces computational overhead, as it avoids recomputing the key and value tensors for all prior tokens at every decoding step, resulting in faster inference.  However, the KV cache may become a memory bottleneck when handling long contexts or large batch sizes (see Appendix~\ref{append::kvmemory}). Compressing the KV cache---through methods such as quantization---offers a natural solution to this challenge.

\paragraph{Quantization.} 
In practice, quantization is often applied to a group of numbers $x_1,\cdots,x_n$ simultaneously.  In the context of KV cache quantization, these numbers can be elements of the KV cache for the same token across hidden dimensions (per-token quantization) or elements from multiple tokens within the same hidden dimension (per-channel quantization). Quantization begins by determining $m=\min\{x_i\}_{i=1}^n$ and $M = \max\{x_i\}_{i=1}^n$, and then maps each number to a $b$-bit integer using: 
\begin{align*}
    &\mathsf{qtz}(x_i) = \left\lfloor \frac{x_i - m}{\Delta}\right\rceil \text{ with } \Delta = \frac{M-m}{2^b - 1}.
\end{align*}
Here $m$ and $\Delta$ are referred to as the zero-point and scaling factor, respectively. The operator $\lfloor \cdot \rceil$ rounds to the nearest integer. After quantization, the $b$-bit integer falls within the range $[0,2^b-1]$. The quantized values, along with the zero-point and scaling factor, are stored to enable dequantization: $\mathsf{deq}(\bar{x}_i) = \bar{x}_i \times \Delta + m$.

\section{Main Results}

\subsection{Motivation and Problem Formulation}
\label{sec::motivation}

\paragraph{Motivation.} We first recall how existing methods \citep[e.g.,][]{liu2024kivi,kang2024gear,hooper2024kvquant} quantize the KV cache. When a new token produces its key tensor, it is not immediately quantized but instead stored in a residual buffer. Once this buffer accumulates $R$ key tensors, they are quantized together. This process involves grouping the key tensors and computing a zero-point and scaling factor for each group.
Existing methods explore different strategies to reduce quantization error, such as per-channel grouping and pre-RoPE quantization. Despite these differences, they share the same fundamental goal: minimizing the difference between the (de)quantized and original key tensors: $\| \bk - \mathsf{deq}(\bk^{\text{qtz}})\|_2$ (with the same principle applied to value tensors). Based on this criterion, we refer to them as \emph{compression-based quantization methods}.

Compression-based methods overlook the impact of quantization on the attention mechanism itself. Since attention scores are determined by the inner products between query and key vectors, even small perturbations in key tensors can lead to disproportionate shifts in attention outputs. The following theorem formalizes this intuition.
\begin{thm}
\label{thm::att_out_bound}
For a new token with query tensor $\bq_n \in \Reals^{d}$, we can upper bound the output of the attention mechanism when using original keys\&values compared to quantized keys\&values:
\begin{align*}
    &\left\|\mathsf{Attention}(\bq_n, \{\bk_i\}_{i=1}^n, \{\bv_i\}_{i=1}^n)
    - \mathsf{Attention}(\bq_n, \{\bk_i^{\text{qtz}}\}_{i=1}^n , \{\bv_i^{\text{qtz}}\}_{i=1}^n)
    \right\|_2\\
    &\leq \frac{\sum_{i=1}^n \|\bv_i - \mathsf{deq}(\bv_i^{\text{qtz}})\|_2 }{2\sqrt{d}} \sum_{i=1}^n |\bq_n (\bk_i - \mathsf{deq}(\bk_i^{\text{qtz}}))^T| + \sum_{i=1}^n \|\bv_i - \mathsf{deq}(\bv_i^{\text{qtz}}) \|_2.
\end{align*}
\end{thm}
%
%
Theorem~\ref{thm::att_out_bound} indicates that the outputs of the attention mechanism remain unchanged when using quantized tensors, provided two conditions are satisfied: (i) the quantized value tensors closely approximate the original ones, and (ii) the key residuals---the differences between the (de)quantized and original key tensors---are orthogonal to the query tensor. Accordingly, we quantize value tensors on a per-token basis to minimize their approximation errors relative to the original value tensors. For key tensors, we quantize them while ensuring that key residuals remain orthogonal to the query tensors of future tokens.

\paragraph{Problem formulation.} We formalize the above intuition for key tensor quantization as an optimization problem. Given a key tensor $\bk$, our goal is to find its $b$-bit representation $\bk^{\text{qtz}}$ such that the quantization error is minimized, while the residual $\bk-\mathsf{deq}(\bk^{\text{qtz}})$ remains orthogonal to the subspace spanned by the query tensors of future tokens---denoted by $\hat{\bQ}$---to preserve task-relevant information. 
This leads to the following constrained optimization:
\begin{align}
\label{eq::core_opt}
    &\min_{\bk^{\text{qtz}}} \| \bk - \mathsf{deq}(\bk^{\text{qtz}})\|_2, \quad \hat{\bQ} ( \bk - \mathsf{deq}(\bk^{\text{qtz}})) = \bm{0}, \ \bk^{\text{qtz}} \text{ is a vector of b-bit integers}.
\end{align}
We face two challenges to solve this optimization. First, the query tensors of future tokens---and thus the subspace $\hat{\bQ}$---are unavailable when quantizing the current token's key tensor. In Section~\ref{subsec::query_space}, we explore how to estimate $\hat{\bQ}$ without knowing future tokens. Second, the optimization in \eqref{eq::core_opt} is combinatorial and generally intractable. In Section~\ref{sec::key_quant}, we introduce an efficient iterative algorithm to obtain an approximate solution.

\subsection{Exploring Query Subspace}
\label{subsec::query_space}

Our key observation is that query tensors typically lie within a subspace (up to a certain error) whose dimension is significantly smaller than the hidden dimension $d$. 
Importantly, this subspace can be derived directly from the prompt tokens, without requiring access to the response. In other words, even when future tokens are unknown at the time of quantizing the key tensors for the current token, this prompt-derived subspace can still guide the quantization process.

We illustrate this observation using a sequence from the \texttt{math-500} dataset and generate its response with the \texttt{Llama-3.1-8B-Instruct} model. To precisely define what it means for query tensors to lie within a subspace, we introduce the following definition. 
\begin{defn}
For a query tensor $\bq$, we define its deviation from a given subspace $\mathcal{P}$ as $\|\bq - \mathsf{Proj}_{\mathcal{P}}(\bq)\|_2$, where $\mathsf{Proj}_{\mathcal{P}}(\bq)$ denotes the projection of $\bq$ onto $\mathcal{P}$. When the query tensor is normalized, its deviation from any subspace stays within $[0,1]$.
\end{defn}
We apply singular value decomposition (SVD) to query tensors from: (1) all tokens (prompt + response), (2) only prompt tokens, (3) all tokens from a different \texttt{math-500} sequence, and (4) all tokens from a sequence in \texttt{QuALITY} \citep{pang2021quality}, a multiple-choice QA dataset. In each case, we select the top $r$ singular vectors to form the subspace.

\begin{figure*}[t]
    \centering
    \includegraphics[width=0.30\linewidth]{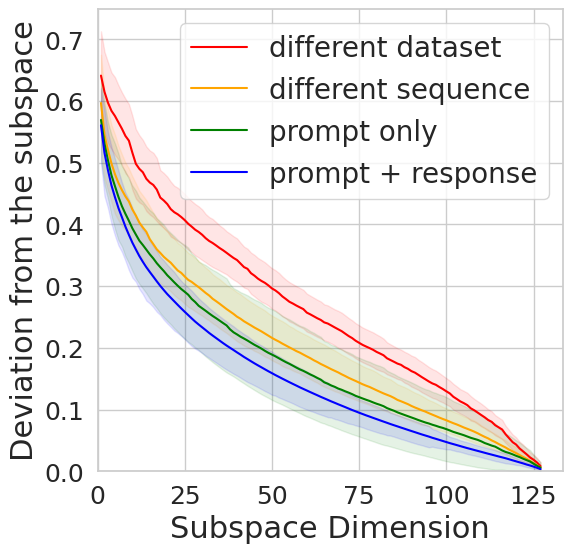}
    \hspace{0.2cm}
    \includegraphics[width=0.64\linewidth]{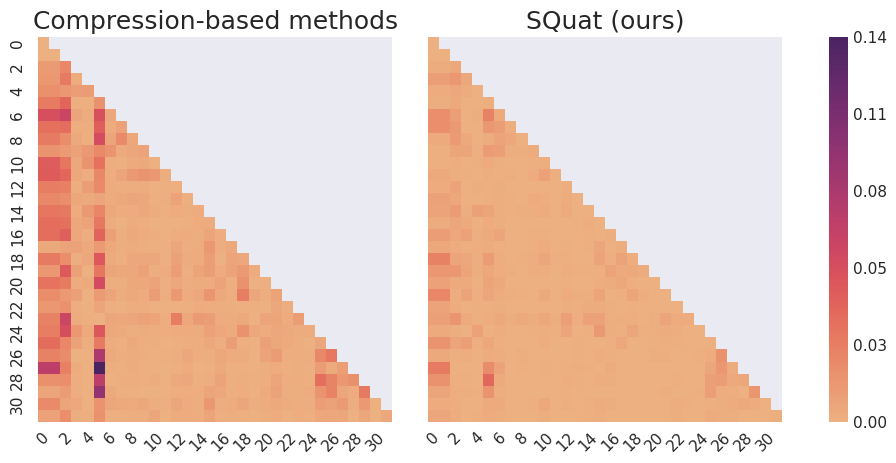}
    \caption{Left: query tensors tend to lie in a low-dimensional subspace. This subspace can be identified using query tensors from prompt tokens, which are available before decoding begins. Middle and Right: the absolute \emph{difference} in attention scores between using original and quantized key tensors. As shown, our method better preserves attention scores. See Appendix~\ref{append::semi_syn} for experimental setup. 
    }
    \label{fig:subspace}
\end{figure*}

Figure~\ref{fig:subspace} (left) illustrates the deviation of (normalized) query tensors of tokens in the \texttt{math-500} sequence from the four subspaces. 
As shown, a 30-dimensional subspace can capture the entire sequence’s query tensors with \textasciitilde$20\%$ deviation error. Moreover, using prompt-only query tensors---available before the decoding phase begins---or query tensors from a different \texttt{math-500} sequence does not increase the deviation error. The deviation increases slightly, however, when the subspace is spanned using a sequence from \texttt{QuALITY}, a different dataset. 
Based on these observations, we apply SVD to the query tensors of all prompt tokens. Then we define the task-relevant query subspace as $\hat{\bQ} \in \Reals^{r \times d}$, constructed from the top $r$ singular vectors scaled by their corresponding singular values.

\subsection{Key Tensors Quantization}
\label{sec::key_quant}

We propose an iterative algorithm to solve \eqref{eq::core_opt}. In each iteration, we selects a coordinate\footnote{In Appendix~\ref{append::block_upd}, we extend our algorithm to do block-by-block quantization of key tensors, rather than element-by-element. In our experiments, each key tensor is partitioned into no more than 4 blocks, leading to at most 4 iterations.} of $\bk$, quantize it, and update the remaining coordinates to ensure the residual is orthogonal to the query subspace as much as possible. The process repeats for the remaining coordinates until all coordinates of $\bk$ are quantized.

Specifically, we use $\hat{\bk}_t \in \Reals^d$ to represent the (de)quantized value of $\bk$ at the $t$-th iteration and set $\hat{\bk}_0 = \bk$. At iteration $t$, the first $t-1$ elements of $\hat{\bk}_t$ remain unchanged from $\hat{\bk}_{t-1}$, as they have already been quantized. We then quantize the $t$-th element and update the remaining $d-t$ elements using \eqref{eq::opt_update} in the following proposition. The hyperparameter $\lambda \in [0,\infty)$ controls the trade-off between making the (de)quantized key close to its original values and enforcing orthogonality of its residuals to the query subspace. When $\lambda = 0$, the algorithm reduces to a compression-based quantization method. For any vector $\bx$, we use $[\bx]_i$ to denote its $i$-th coordinate.

\begin{prop}
\label{prop::upd_key}
Consider the optimization problem
\begin{equation}
\label{eq::gen_upd_opt_s}
\begin{aligned}
    \min_{\hat{\bk}_t\in \Reals^d}~&\|\hat{\bk}_t - \hat{\bk}_{t-1}\|_2^2 + \lambda \|\hat{\bQ} (\hat{\bk}_t - \hat{\bk}_{t-1})\|_2^2, \\ 
    \sto~&[\hat{\bk}_{t}]_i = [\hat{\bk}_{t-1}]_i~~\text{ for } i=1,\cdots,t-1~~\text{and }
    [\hat{\bk}_{t}]_t = \mathsf{deq}(\mathsf{qtz}([\hat{\bk}_{t-1}]_t)).
\end{aligned}
\end{equation}
We denote $\bP_{\text{inv}} \defined (\bI + \lambda \hat{\bQ}^T \hat{\bQ})^{-1}$. Suppose $\bA_{t} \in \Reals^{t\times t}$ is the top-left block of $\bP_{\text{inv}}$ and $\bB_{t} \in \Reals^{(d-t)\times t}$ is the bottom left block of $\bP_{\text{inv}}$. Let $\bh_{t}$ be the last column of $\bA_{t}^{-1}$. Then the optimal solution to~\eqref{eq::gen_upd_opt_s} has its last $d - t$ elements given by 
\begin{align}
\label{eq::opt_update}
[\hat{\bk}_{t-1}]_{t:} + (\mathsf{deq}(\mathsf{qtz}([\hat{\bk}_{t-1}]_t)) - [\hat{\bk}_{t-1}]_t)\bB_{t}\bh_{t}.
\end{align}
\end{prop}
%

The main computational cost in Proposition~\ref{prop::upd_key} arises from computing $\bA_{t}^{-1}$ at each iteration, which has a complexity of $O(\sum_{t=1}^d t^3) = O(d^4)$. The following proposition introduces an iterative algorithm to compute $\bA_{t}^{-1}$. It begins with $\bA_{d}^{-1} = \bI + \lambda \hat{\bQ}^T \hat{\bQ}$ by definition and subsequently derives $\bA_{t}^{-1}$ from $\bA_{t+1}^{-1}$. This way, we reduce the complexity of the iterative algorithm in Proposition~\ref{prop::upd_key} from $O(d^4)$ to $O(d^3)$. 

\begin{prop}
\label{prop::rem_one}
Recall from Proposition~\ref{prop::upd_key} that $\bA_{t+1} \in \Reals^{(t+1) \times (t+1)}$ is the top-left block of $\bP_{\text{inv}}$. For $t=d-1,\cdots,1$,  if we write $\bA_{t+1}^{-1}= \begin{bmatrix}
    \bar{\bA}_{t+1} & \ba_{t+1}^T\\
    \ba_{t+1} & \bar{a}_{t+1}
\end{bmatrix}$, 
where $\bar{\bA}_{t+1} \in \Reals^{t\times t}$, then we have
\begin{align}
    \bA_{t}^{-1} 
    = \bar{\bA}_{t+1} - \frac{1}{\bar{a}_{t+1}} \ba_{t+1}^T \ba_{t+1}.
\end{align}
\end{prop}
%
%
We show how our method preserves attention scores compared to compression-based methods in Figure~\ref{fig:subspace} (middle and right) and defer a more comprehensive evaluation to the next section. A pseudo-code of our algorithm is provided in Algorithm~\ref{alg::main_alg}, with an illustration shown in Figure~\ref{fig:our_algo}.

\section{Numerical Experiments}
\label{sec::exp}

\definecolor{lightblue}{rgb}{0.88, 0.92, 1}
\begin{table*}[t]
\small
\centering
\resizebox{0.92\textwidth}{!}{
\renewcommand{\arraystretch}{1.25}
\begin{tabular}{lcccccccc}
\toprule
\textbf{Method} & \textbf{KV size} & \textbf{GSM8k} & \textbf{MMLU\_Pro\_Math} & \textbf{MMLU\_Pro\_Law} & \textbf{IFEval} & \textbf{GPQA} & \textbf{BBH} & \textbf{Average} \\
\midrule

\multicolumn{9}{c}{\textbf{\texttt{Llama-2-7B-hf}}} \\ 
FP16 & 100\% & 14.03\% & 7.25\% & 16.17\% & 27.47\% & 11.61\% & 39.53\% & 20.87\% \\ 
\midrule
\rowcolor{lightblue}
KIVI & 19.7\% & 11.98\% & \underline{8.51\%} & 12.53\% & 26.29\% & 9.60\% & \underline{35.92\%} & 18.86\% \\ 
GEAR & 22.5\% & 11.60\% & \textbf{9.25\%} & \underline{13.72\%} & \textbf{27.94\%} & 9.60\% & 35.77\% & 19.28\% \\ 
\rowcolor{lightblue}
ZipCache & 21.6\% & 6.37\% & 7.48\% & \textbf{15.80\%} & 25.71\% & 8.48\% & 24.73\% & 15.39\% \\ 
\textbf{SQuat} & 19.7\% & \textbf{13.42\%} & 7.62\% & \underline{15.35\%} & \underline{27.44\%} & \underline{9.82\%} & 35.76\% & \textbf{19.59\%} \\ 
\rowcolor{lightblue}
\textbf{SQuat\textsuperscript{pre}} & 19.7\% & \underline{12.21\%} & 6.66\% & 13.72\% & 25.62\% & \textbf{11.83\%} & \textbf{36.85\%} & \underline{19.34\%} \\ 
\bottomrule 

\multicolumn{9}{c}{\textbf{\texttt{Llama-3.1-8B-Instruct}}} \\ 
FP16 & 100\% & 77.56\% & 39.38\% & 26.70\% & 52.96\% & 16.74\% & 71.06\% & 50.27\% \\ 
\midrule
\rowcolor{lightblue}
KIVI & 21.4\% & 71.49\% & 30.72\% & 27.25\% & 50.80\% & 13.39\% & 59.62\% & 44.86\% \\ 
GEAR & 25.8\% & 69.37\% & 30.35\% & \textbf{27.79\%} & \underline{54.32\%} & 15.40\% & 57.70\% & 45.17\% \\ 
\rowcolor{lightblue}
ZipCache & 22.1\% & 72.02\% & 32.72\% & \underline{27.43\%} & 51.83\% & 11.16\% & 55.32\% & 44.08\% \\ 
\textbf{SQuat} & 21.4\% & \textbf{72.71\%} & \textbf{33.83\%} & 27.07\% & 54.14\% & \underline{16.96\%} & \underline{60.59\%} & \underline{46.97\%} \\ 
\rowcolor{lightblue}
\textbf{SQuat\textsuperscript{pre}} & 21.4\% & \underline{72.56\%} & \underline{32.94\%} & 27.34\% & \textbf{55.91\%} & \textbf{17.86\%} & \textbf{62.85\%} & \textbf{47.86\%} \\ 
\bottomrule 

\multicolumn{9}{c}{\textbf{\texttt{Mistral-7B-Instruct-v0.3}}} \\ 
FP16 & 100\% & 50.49\% & 23.09\% & 24.34\% & 54.21\% & 18.30\% & 56.23\% & 40.59\% \\ 
\midrule
\rowcolor{lightblue}
KIVI & 20.7\% & 42.61\% & 20.21\% & 23.07\% & \textbf{54.01\%} & \underline{18.53\%} & 47.77\% & 36.91\% \\ 
\textbf{SQuat} & 20.7\% & \textbf{45.26\%} & \underline{21.24\%} & \underline{23.52\%} & \underline{53.82\%} & \textbf{19.87\%} & \underline{50.27\%} & \textbf{38.32\%} \\ 
\rowcolor{lightblue}
\textbf{SQuat\textsuperscript{pre}} & 20.7\% & \underline{44.50\%} & \textbf{22.95\%} & \textbf{23.98\%} & 52.28\% & \underline{18.53\%} & \textbf{50.79\%} & \underline{37.91\%} \\ 
\bottomrule 
\end{tabular}
}
\caption{We compare SQuat and SQuat\textsuperscript{pre} against existing tuning-free KV cache quantization baselines and FP16 (no quantization) across multiple tasks from LM-Eval.
KV size is the average ratio of the compressed KV cache to its FP16 counterpart. For each task, the best result is shown in bold, and the second-best is underlined.}
\label{table:exp_lmeval}
\end{table*}

We evaluate SQuat and SQuat\textsuperscript{pre} for 2-bit quantization of the KV cache using four LLMs across 14 benchmark tasks, comparing them with four baseline quantization methods. 
The key difference between SQuat and SQuat\textsuperscript{pre} is that SQuat\textsuperscript{pre} quantizes and stores the KV cache \emph{before applying RoPE} \citep{su2024roformer}, as recommended by \citet{hooper2024kvquant}. Because SQuat\textsuperscript{pre} applies positional embeddings on-the-fly after dequantization, it operates slightly slower than SQuat.

\paragraph{Baselines and LLMs.} We consider FP16 (no quantization) and existing KV cache quantization algorithms---KIVI \citep{liu2024kivi}, GEAR \citep{kang2024gear}, ZipCache \citep{he2025zipcache}---which, like our method, do not require fine-tuning or a calibration dataset. We evaluate these methods on four open-source LLMs: \texttt{Llama-2-7B},
\texttt{Llama-3.1-8B-Instruct},
\texttt{Mistral-7B-Instruct-v0.3},
\texttt{DeepSeek-R1-Distill-Llama-8B}. We assess performance across several reasoning and long-context benchmarks, and report the KV size as the average percentage of the compressed cache relative to the FP16 cache at the end of generation. We present the experimental setup, including the choice of hyper-parameters for SQuat, SQuat\textsuperscript{pre}, and other baseline methods in Appendix~\ref{append::setup}.

\paragraph{Reasoning tasks.} We evaluate SQuat and baseline methods on a range of reasoning tasks from \texttt{LM-Eval} \citep{evalharness}, including \texttt{GSM8k}, \texttt{MMLU\_Pro\_Math}, \texttt{MMLU\_Pro\_Law}, 
\texttt{IFEval}, \texttt{GPQA}, and \texttt{BBH}. These benchmarks are widely adopted in prior work on KV cache compression and are known for their difficulty, often requiring multi-step reasoning (e.g., chain-of-thought) before arriving at the final answer. In such settings, KV cache quantization plays a critical role by substantially reducing memory consumption. We present the results in Table~\ref{table:exp_lmeval}. As shown, SQuat achieves higher average scores across all models. Additionally, on GSM8k, GPQA, and BBH---three particularly challenging reasoning tasks---our method consistently outperforms all baselines by a significant margin.

\paragraph{LongBench tasks.}
We evaluate SQuat and baseline methods on long-context benchmarks, following the recommendations of \citet{liu2024kivi} and using the same set of tasks (\texttt{Qasper}, \texttt{QMSum}, \texttt{MultiNews}, \texttt{TREC}, \texttt{TriviaQA}, \texttt{SAMSum}, \texttt{LCC}, and \texttt{RepoBench-P}) from LongBench \citep{bai2023longbench}. These tasks evaluate the ability of LLMs to process and reason over long contexts, such as document-level question answering and summarization, and require deep understanding across diverse real-world scenarios. As shown in Table~\ref{table:exp_long_bench}, SQuat achieves higher average scores across all models than other quantization baselines and are nearly lossless compared with FP16.

\begin{table*}[t]
\small
\centering
\resizebox{0.98\textwidth}{!}{
\renewcommand{\arraystretch}{1.25}
\begin{tabular}{lcccccccccc}
\toprule
\textbf{Method} & \textbf{KV size} & \textbf{Qasper} & \textbf{QMSum} & \textbf{MultiNews} & \textbf{TREC} & \textbf{TriviaQA} & \textbf{SAMSum} & \textbf{LCC} & \textbf{RepoBench-P} & \textbf{Average} \\
\midrule
\multicolumn{11}{c}{\textbf{\texttt{Llama-2-7B-hf}}} \\ 
FP16 & 100\% & 9.61\%  & 21.15\%  & 3.51\%  & 66.00\%  & 87.72\%  & 41.53\%  & 66.66\%  & 59.81\%  & 44.50\% \\ 
\midrule
\rowcolor{lightblue}
KIVI & 19.1\% & 9.43\%  & 20.71\%  & 0.92\%  & \textbf{66.00\%}  & 87.42\%  & \textbf{42.69\%}  & \underline{66.47\%}  & \textbf{59.83\%}  & 44.18\% \\ 
GEAR & 20.7\% & 9.77\%  & 20.47\%  & 0.96\%  & \textbf{66.00\%}  & 87.42\%  & 42.42\%  & \textbf{66.82\%}  & 59.48\%  & 44.17\% \\ 
\rowcolor{lightblue}
ZipCache & 19.9\% & 9.22\%  & 19.65\%  & \textbf{3.97\%}  & 64.50\%  & \underline{87.48\%}  & 40.21\%  & 63.44\%  & 55.99\%  & 43.06\%\\ 
\textbf{SQuat} & 19.1\% & \textbf{10.32\%} & \textbf{20.97\%} & \underline{3.48\%}  & \textbf{66.00\%}  & 87.28\%  & 42.27\%  & 66.36\%  & \underline{59.59\%}  & \textbf{44.53\%}\\ 
\rowcolor{lightblue}
\textbf{SQuat\textsuperscript{pre}} & 19.1\% & \underline{9.94\%}  & \textbf{20.97\%} & 2.44\%  & 65.50\%  & \textbf{88.09\%}  & \underline{42.44\%}  & 66.29\%  & 59.27\%  & \underline{44.37\%}\\ 
\bottomrule 
\multicolumn{11}{c}{\textbf{\texttt{Llama-3.1-8B-Instruct}}} \\
FP16 & 100\% & 45.00\% & 23.40\% & 27.21\% & 69.50\% & 91.20\% & 44.01\% & 63.45\% & 55.25\% & 52.38\%\\ 
\midrule
\rowcolor{lightblue}
KIVI & 19.6\% & 44.03\% & \underline{23.86\%} & \textbf{26.79\%} & \textbf{69.50\%} & 91.85\% & 43.47\% & 62.45\% & 53.04\% & 51.87\%\\ 
GEAR & 21.4\% & 43.13\% & 23.26\% & \textbf{26.79\%} & 69.00\% & 90.74\% & 43.21\% & 61.82\% & 52.87\% & 51.35\%\\
\rowcolor{lightblue}
ZipCache & 21.4\% & 42.87\% & 23.24\% & 26.61\% & \textbf{69.50\%} & 91.66\% & 43.43\% & 62.64\% & 51.86\% & 51.48\%\\ 
\textbf{SQuat} & 19.6\% & \underline{44.22\%} & 23.55\% & 26.78\% & \textbf{69.50\%} & \textbf{92.19\%} & \textbf{44.19\%} & \underline{62.94\%} & \underline{53.09\%} & \underline{52.06\%}\\ 
\rowcolor{lightblue}
\textbf{SQuat\textsuperscript{pre}} & 19.6\% & \textbf{44.25\%} & \textbf{23.87\%} & 26.78\% & \textbf{69.50\%} & \underline{92.02\%} & \underline{43.88\%} & \textbf{63.00\%} & \textbf{53.53\%} & \textbf{52.10\%}\\ 
\bottomrule 
\multicolumn{11}{c}{\textbf{\texttt{Mistral-7B-Instruct-v0.3}}} \\ 
FP16 & 100\% & 40.54\% & 24.34\% & 27.83\% & 74.00\% & 88.39\% & 47.74\% & 58.39\% & 56.89\% & 52.27\%\\ 
\midrule
\rowcolor{lightblue}
KIVI & 19.2\% & 38.87\% & \underline{23.90\%} & 26.94\% & \textbf{74.00\%} & \underline{88.34\%} & \textbf{47.74\%} & 57.64\% & \underline{55.81\%} & \underline{51.66\%}\\ 
\textbf{SQuat} & 19.2\% & \textbf{39.57\%} & \textbf{23.92\%} & \underline{27.01\%} & \textbf{74.00\%} & \textbf{88.89\%} & \underline{46.94\%} & \textbf{57.82\%} & 55.77\% & \textbf{51.74\%}\\ 
\rowcolor{lightblue}
\textbf{SQuat\textsuperscript{pre}} & 19.2\% & \underline{39.04\%} & 23.80\% & \textbf{27.31\%} & \textbf{74.00\%} & 87.94\% & 46.93\% & \underline{57.67\%} & \textbf{56.42\%} & 51.64\%\\ 
\bottomrule 
\end{tabular}
}
\caption{Comparison of different KV cache quantization methods on tasks from LongBench. For each task, the best result is shown in bold, and the second-best is underlined.}
\label{table:exp_long_bench}
\end{table*}

\paragraph{Reasoning model.}

\begin{figure*}[t]
    \centering
    \includegraphics[width=0.71\linewidth]{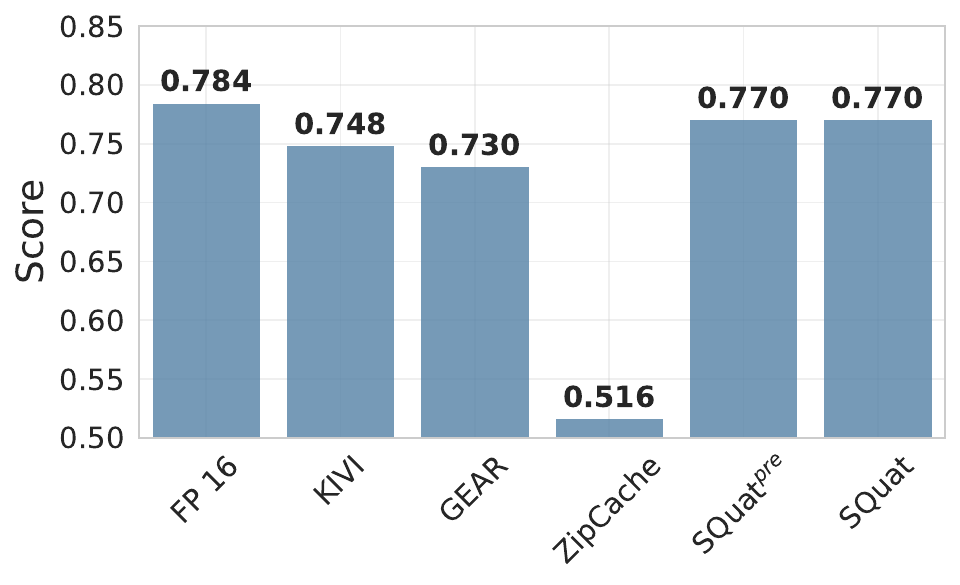}
    \vspace{-1.2em}
    \caption{Experiments using a reasoning model (\texttt{DeepSeek-R1-Distill-Llama-8B}) on the \texttt{math-500} dataset. We compare SQuat and SQuat\textsuperscript{pre} with different KV cache quantization baselines and FP16 (no quantization).
    }
    \label{fig:math500}
\end{figure*}

We evaluate \texttt{DeepSeek-R1-Distill-Llama-8B} on the \texttt{math-500} dataset \citep{lightman2023lets}. We set the maximum number of new tokens to 8k to ensure the model has sufficient window for multi-stage reasoning. This setup highlights a key use case for KV cache quantization: modern reasoning models often require long responses to ``think'' through problems---e.g., by generating chain-of-thought reasoning---before arriving at a final answer. In such scenarios, the KV cache becomes a memory bottleneck, and quantization proves especially helpful when compute resources are limited. As shown in Figure~\ref{fig:math500}, both SQuat and SQuat\textsuperscript{pre} outperform all existing KV cache quantization baselines.

\paragraph{Efficiency comparisons.}

\begin{figure*}[t]
    \centering
    \includegraphics[width=0.32\linewidth]{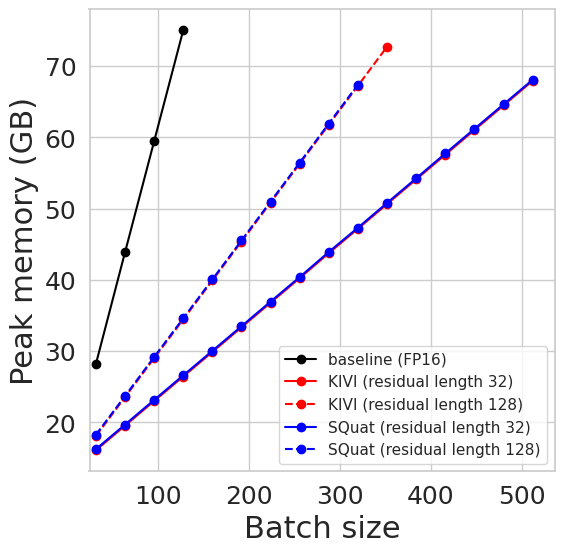}
    \includegraphics[width=0.32\linewidth]{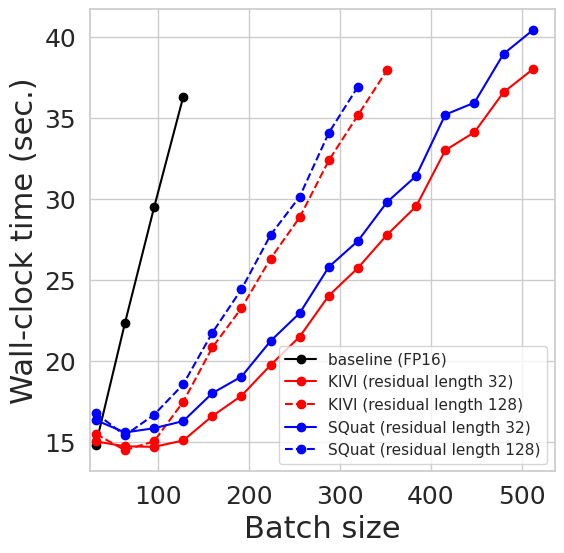}
    \includegraphics[width=0.32\linewidth]{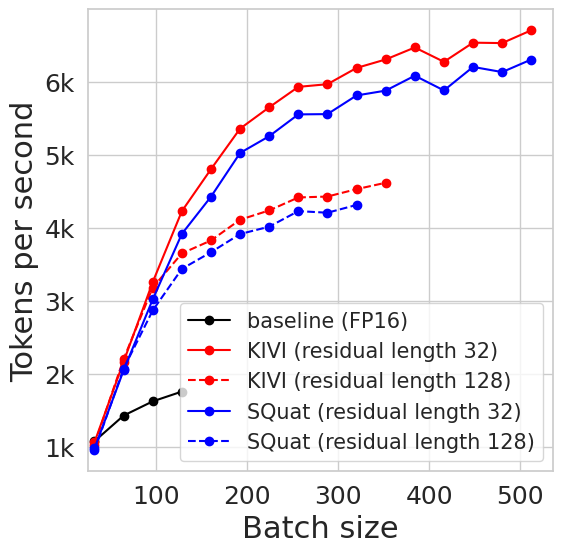}
    \caption{We evaluate SQuat's efficiency by measuring peak memory usage (left), latency (middle), and throughput (right), comparing it against KIVI \citep{liu2024kivi} and an FP16 baseline. As shown, SQuat requires the same peak GPU memory as KIVI. In terms of throughput, SQuat achieves a $2.45\times \sim 3.60\times$ improvement over FP16, which is comparable to KIVI’s $2.63\times \sim 3.82\times$ improvement.
    }
    \label{fig:mem_throughput}
\end{figure*}

We apply SQuat to quantize the KV cache to 2-bit and evaluate its efficiency in terms of memory usage, latency, and throughput on the \texttt{Llama-2-7B} model. Empirically, GEAR and ZipCache run much slower than KIVI \citep{liu2024kivi} and SQuat, so we compare SQuat only against KIVI and an FP16 baseline (without quantization) in Figure~\ref{fig:mem_throughput}.
To ensure a fair comparison, we adopt the same experimental setup as KIVI, which synthesizes workloads based on \citet{sharegpt}. We evaluate two residual buffer lengths (i.e., the number of most recent tokens whose KV cache is stored in full precision), 32 and 128, for both our algorithm and KIVI. We gradually increase the batch size until the GPU runs out of memory. All experiments are conducted on a single Nvidia H100 GPU (80GB). 
As shown in Figure~\ref{fig:mem_throughput}, SQuat reduces peak GPU memory by $\mathbf{2.17\times} \sim \mathbf{2.82\times}$. This memory efficiency enables up to a $4\times$ increase in batch size compared to the FP16 baseline, leading to a throughput improvement of $\mathbf{2.45\times \sim 3.60 \times}$, which is comparable to KIVI’s throughput improvement of $\mathbf{2.63\times \sim 3.82 \times}$.

\section{Conclusion, Future work, and Limitations}

The past years have seen rapid advancements in techniques for developing LLMs. As SOTA models continue to grow in size to achieve higher performance \citep{zhao2023survey}, ensuring accessibility to these techniques---especially for those with limited computational resources---has become increasingly important. This paper responds to this call by exploring methods to make LLM inference-time decoding more efficient. We focus on a specific aspect of this challenge: KV cache quantization. Unlike existing compression-based algorithms, our method quantizes key tensors with the objective of preserving their inner product with future tokens' query tensors. This design choice ensures that quantization errors minimally affect model outputs. We establish rigorous theoretical foundations for our algorithm and hope our work inspires new research that applies theoretical insights to practical LLM development, promoting efficient and accessible inference techniques.

There are several promising directions for further exploration. First, some recent LLMs do not store the KV cache directly. For instance, multi-head latent attention \citep{liu2024deepseek} caches compressed latent vectors and applies projection matrices to generate key and value tensors during inference. Investigating whether these latent vectors can be further quantized, and understanding the impact of quantization errors on model performance, is an interesting avenue for research. From a theoretical standpoint, it is also valuable to examine how reducing KV cache size---through quantization or pruning---affects latency, memory usage, and throughput. Identifying the optimal trade-offs between compression rate and model performance for specific tasks could offer insights into more efficient model deployment strategies.

\clearpage
\section*{Acknowledgment}
We would like to thank Shivchander Sudalairaj, Abhishek Bhandwaldar, Mustafa Eyceoz, Aldo Pareja from the Red Hat AI Innovation Team for their help in setting up some of the evaluation benchmarks. 

\bibliography{colm2025_conference}
\bibliographystyle{colm2025_conference}

\appendix

\section{Omitted Proofs}

\subsection{Proof of Theorem~\ref{thm::att_out_bound}}
\label{subsec::att_output}

\begin{proof}
We denote $\bk_i^{\text{deq}} \defined \mathsf{deq}(\bk_i^{\text{qtz}})$ and $\bv_i^{\text{deq}} \defined \mathsf{deq}(\bv_i^{\text{qtz}})$ for $i=1,\cdots,n$. We introduce two vectors:
\begin{align*}
\bw = \left[\frac{\langle \bq_n, \bk_1 \rangle}{\sqrt{d}}, \cdots, \frac{\langle \bq_n, \bk_n\rangle}{\sqrt{d}}\right],
\quad 
\bw^{\text{deq}} = \left[\frac{\langle \bq_n, \bk_1^{\text{deq}}\rangle}{\sqrt{d}}, \cdots, \frac{\langle\bq_n, \bk_n^{\text{deq}}\rangle}{\sqrt{d}}\right].
\end{align*}
Let $\bV, \bV^{\text{deq}} \in \Reals^{n\times d}$, where the rows consist of $\bv_i$ and $\bv_i^{\text{deq}}$, respectively. Then we have
\begin{align*}
    &\left\|\mathsf{Attention}(\bq_n, \{\bk_i\}_{i=1}^n, \{\bv_i\}_{i=1}^n)
    - \mathsf{Attention}(\bq_n, \{\bk_i^{\text{qtz}}\}_{i=1}^n , \{\bv_i^{\text{qtz}}\}_{i=1}^n)
    \right\|_2\\
    &= \| \mathsf{softmax}(\bw)\bV - \mathsf{softmax}(\bw^{\text{deq}})\bV^{\text{deq}}\|_2\\
    &= \| \mathsf{softmax}(\bw)(\bV - \bV^{\text{deq}})
    + (\mathsf{softmax}(\bw)
    -\mathsf{softmax}(\bw^{\text{deq}}))\bV^{\text{deq}}\|_2\\
    &\leq \| \mathsf{softmax}(\bw)(\bV - \bV^{\text{deq}})\|_2
    +\| (\mathsf{softmax}(\bw)
    -\mathsf{softmax}(\bw^{\text{deq}}))\bV^{\text{deq}}\|_2.
\end{align*}
Note that 
\begin{align*}
    \| \mathsf{softmax}(\bw)(\bV - \bV^{\text{deq}})\|_2
    \leq \sum_{i=1}^n \|\bv_i - \bv_i^{\text{deq}} \|_2.
\end{align*}
Similarly, we have 
\begin{align*}
    \| (\mathsf{softmax}(\bw)
    -\mathsf{softmax}(\bw^{\text{deq}}))\bV^{\text{deq}}\|_2
    \leq \|\mathsf{softmax}(\bw)
    -\mathsf{softmax}(\bw^{\text{deq}}) \|_2 \|\bV^{\text{deq}}\|_F.
\end{align*}
Note that $\mathsf{softmax}$ is $1/2$-Lipschitz continuous \citep[see Appendix~A.4 in][]{alghamdi2022beyond}. Therefore,
\begin{align*}
    &\|\mathsf{softmax}(\bw)
    -\mathsf{softmax}(\bw^{\text{deq}}) \|_2
    \leq \frac{1}{2} \| \bw - \bw^{\text{deq}}\|_2\\
    &\leq \frac{1}{2\sqrt{d}} \sqrt{\langle \bq_n, (\bk_1 - \bk_1^{\text{deq}}) \rangle^2 + \cdots + \langle \bq_n, (\bk_n - \bk_n^{\text{deq}}) \rangle^2}\\
    &\leq \frac{1}{2\sqrt{d}} \sum_{i=1}^n |\bq_n (\bk_i - \bk_i^{\text{deq}})^T|.
\end{align*}
\end{proof}

\subsection{Block-Wise Quantization}
\label{append::block_upd}

The algorithm presented in the main body quantizes the key vectors element by element. In this section, we extend it to perform block-wise quantization, improving efficiency by quantizing multiple elements of the key vectors simultaneously while updating the remaining elements. As special cases of this approach, we prove Proposition~\ref{prop::upd_key} and Proposition~\ref{prop::rem_one}.

Before diving into the details, we first recall some standard results from linear algebra that will be used in our proofs. 
\begin{lem}
\label{lem::inv_rm}
Consider a block matrix:
\begin{align*}
\bM = 
\begin{bmatrix}
    \bA & \bB\\
    \bC & \bD
\end{bmatrix},
\end{align*}
where $\bA$ and $\bD$ are square matrices. We denote $\bM/\bD \defined \bA - \bB \bD^{-1} \bC$ and
$\bM/\bA \defined \bD - \bC \bA^{-1} \bB$. Suppose $\bD$ and $\bM/\bD$ are both invertible. Then
\begin{align*}
\bM^{-1}
= 
\begin{bmatrix}
(\bM/\bD)^{-1} & -(\bM/\bD)^{-1}\bB \bD^{-1} \\
-\bD^{-1}\bC (\bM/\bD)^{-1} & \bD^{-1} + \bD^{-1} \bC (\bM/\bD)^{-1} \bB \bD^{-1}
\end{bmatrix}.
\end{align*}
Similarly, suppose $\bA$ and $\bM/\bA$ are both invertible. Then
\begin{align*}
\bM^{-1}
=
\begin{bmatrix}
\bA^{-1} + \bA^{-1} \bB(\bM/\bA)^{-1} \bC \bA^{-1} & -\bA^{-1} \bB(\bM/\bA)^{-1}\\
-(\bM/\bA)^{-1} \bC \bA^{-1} & (\bM/\bA)^{-1}
\end{bmatrix}.
\end{align*}
\end{lem}
Using this lemma, we immediately obtain the following corollary, which is also a standard result in linear algebra.
\begin{cor}
\label{cor::rem}
We denote
\begin{align*}
\bM = 
\begin{bmatrix}
    \bA & \bB\\
    \bC & \bD
\end{bmatrix},
\quad
\bM^{-1}
= 
\begin{bmatrix}
    \bP & \bQ\\
    \bR & \bS
\end{bmatrix}.
\end{align*}
Suppose $\bA$ and $\bM/\bA$ are both invertible. Then we have $\bA^{-1} = \bP - \bQ \bS^{-1} \bR$. 
\end{cor}
\begin{proof}
Lemma~\ref{lem::inv_rm} implies that
\begin{align*}
&\bP = \bA^{-1} + \bA^{-1} \bB(\bM/\bA)^{-1} \bC \bA^{-1},\\
&\bQ = -\bA^{-1} \bB(\bM/\bA)^{-1},\\
&\bR = -(\bM/\bA)^{-1} \bC \bA^{-1},\\
&\bS = (\bM/\bA)^{-1}.
\end{align*}
Substituting the last three equations into the first one yields
\begin{align*}
\bA^{-1} 
= \bP - \bA^{-1} \bB(\bM/\bA)^{-1} \bC \bA^{-1}
= \bP - \bQ \bS^{-1} \bR.
\end{align*}
\end{proof}

Now, we consider quantizing a block of $g$ elements per iteration in a sequential manner. Suppose the hidden dimension $d$ of the key tensors is divisible by $g$. At iteration $t$, the first $(t-1)g$ elements have already been quantized and are kept fixed. During this step, we quantize the elements indexed from $(t-1)g+1$ to $tg$ and update the rest of the elements by using the following lemma.

\begin{lem}
\label{lem::block_quan}
Consider the optimization problem
\begin{equation}
\label{eq::gen_upd_opt}
\begin{aligned}
    \min_{\hat{\bk}_t\in \Reals^d}~&\|\hat{\bk}_t - \hat{\bk}_{t-1}\|_2^2 + \lambda \|\hat{\bQ} (\hat{\bk}_t - \hat{\bk}_{t-1})\|_2^2, \\ 
    \sto~&[\hat{\bk}_{t}]_i = [\hat{\bk}_{t-1}]_i\quad\text{ for } i=1,\cdots,(t-1)g\\
    &[\hat{\bk}_{t}]_i = \mathsf{deq}(\mathsf{qtz}([\hat{\bk}_{t-1}]_i))\quad \text{ for } i = (t-1)g+1,\cdots,tg. 
\end{aligned}
\end{equation}
We denote 
\begin{align*}
    \bP_{\text{inv}} 
    \defined (\bI + \lambda \hat{\bQ}^T \hat{\bQ})^{-1}
    = 
    \begin{bmatrix}
    \bA_{t} & \bB_{t}^T\\
    \bB_{t} & \bC_{t}
    \end{bmatrix},
\end{align*}
where $\bA_{t} \in \Reals^{tg\times tg}$ is the top-left block of $\bP_{\text{inv}}$. Let $\bH_{t}$ be the last $g$ columns of $\bA_{t}^{-1}$. 
The optimization in \eqref{eq::gen_upd_opt} has a closed-form optimal solution:
\begin{align*}
[\hat{\bk}_t]_i
= 
\begin{cases}
[\hat{\bk}_{t-1}]_{i} \quad i = 1,\cdots,(t-1)g\\
 \mathsf{deq}(\mathsf{qtz}([\hat{\bk}_{t-1}]_i)) \quad i=(t-1)g+1,\cdots,tg
\end{cases}
\end{align*}
and the remaining elements of $\hat{\bk}_t$ are given by 
\begin{align}
\label{eq::update_block}
[\hat{\bk}_{t-1}]_{tg:} + \bB_{t}\bH_{t} \bd, 
\end{align}
where $\bd \in \Reals^g$ has its $i$-th element corresponding to the $(t-1)g+i$-th element of $\mathsf{deq}(\mathsf{qtz}(\hat{\bk}_{t-1})) - \hat{\bk}_{t-1}$.
\end{lem}
\begin{proof}
We denote $\bdelta = \hat{\bk}_t - \hat{\bk}_{t-1}$ and let $\bd \in \Reals^g$ has its $i$-th element corresponding to the $(t-1)g+i$-th element of $\mathsf{deq}(\mathsf{qtz}(\hat{\bk}_{t-1})) - \hat{\bk}_{t-1}$. Moreover, we denote
\begin{align*}
    \bT \defined [\bI, \bm{0}],
    \quad
    \bb \defined \begin{bmatrix}
        \bm{0}\\
        \bd
    \end{bmatrix},
\end{align*}
where the identity matrix has dimension $tg \times tg$ and $\bb \in \Reals^{tg}$. The optimization problem in \eqref{eq::gen_upd_opt} can be rewritten as
\begin{align*}
    \min_{\bdelta \in \Reals^d} \bdelta^T \bdelta + \lambda \bdelta^T \hat{\bQ}^T \hat{\bQ} \bdelta , \quad \bT \bdelta = \bb.
\end{align*}
This expression can be simplified to 
\begin{align}
\label{eq::opt_simp}
    \min_{\bdelta \in \Reals^d} \bdelta^T \bP \bdelta, \quad \bT \bdelta = \bb,
\end{align}
where we denote $\bP \defined \bI + \lambda \hat{\bQ}^T \hat{\bQ}$. Note that $\bP$ is positive definite so it is invertible. We denote $\bP_{\text{inv}} \defined \bP^{-1}$. The Lagrangian of \eqref{eq::opt_simp} is:
\begin{align*}
    L(\bdelta, \blambda)
    = \bdelta^T \bP \bdelta - \blambda^T (\bT\bdelta - \bb).
\end{align*}
The KKT conditions give
\begin{align}
    2\bP\bdelta &= \bT^T \blambda, \label{eq::delta_lamb}\\
    \bT\bdelta &= \bb.
\end{align}
Hence, we have
\begin{align*}
    &\bdelta = \frac{1}{2}\bP_{\text{inv}} \bT^T \blambda,\\
    &\frac{1}{2}\bT \bP_{\text{inv}} \bT^T \blambda = \bb.
\end{align*}
Suppose $\bT \bP_{\text{inv}} \bT^T$ is invertible. Then we have
\begin{align*}
    \blambda = 2(\bT \bP_{\text{inv}} \bT^T)^{-1}\bb.
\end{align*}
Substituting the above equation into \eqref{eq::delta_lamb} leads to
\begin{align}
\label{eq::delta_1}
    \bdelta = \bP_{\text{inv}}\bT^T (\bT \bP_{\text{inv}} \bT^T)^{-1}\bb.
\end{align}
Recall that $\bT = [\bI, \bm{0}]$ and $\bP_{\text{inv}} = 
\begin{bmatrix}
\bA_{t} & \bB_{t}^T\\
\bB_{t} & \bC_{t}
\end{bmatrix}$.
Hence, we have
\begin{align*}
    \bP_{\text{inv}}\bT^T
    = 
    \begin{bmatrix}
        \bA_{t}\\
        \bB_{t}
    \end{bmatrix}
    ,
    \quad
    \bT \bP_{\text{inv}} \bT^T
    = \bA_{t}.
\end{align*}
As a result, we have 
\begin{align*}
    \bP_{\text{inv}}\bT^T (\bT \bP_{\text{inv}} \bT^T)^{-1}
    = 
    \begin{bmatrix}
        \bA_{t}\\
        \bB_{t}
    \end{bmatrix}
    (\bA_{t})^{-1}
    = 
    \begin{bmatrix}
        \bI \\
        \bB_{t}(\bA_{t})^{-1}
    \end{bmatrix}.
\end{align*}
Substituting the above equation and $\bb = \begin{bmatrix} \bm{0}\\ \bd \end{bmatrix}$ into \eqref{eq::delta_1} leads to 
\begin{align*}
    \bdelta 
    = 
    \begin{bmatrix}
        \bm{0} \\
        \bd\\
        \bB_{t}(\bA_{t})^{-1} \bb
    \end{bmatrix}.
\end{align*}
Recall that $\bH_{t}$ is the last $g$ columns of $(\bA_{t})^{-1}$. Then 
\begin{align*}
    \bdelta 
    = 
    \begin{bmatrix}
        \bm{0} \\
        \bd\\
        \bB_{t}\bH_{t} \bd
    \end{bmatrix}.
\end{align*}
\end{proof}

Now we are in position to prove Proposition~\ref{prop::upd_key}. 
\begin{proof}[Proof of Proposition~\ref{prop::upd_key}]
Proposition~\ref{prop::upd_key} can be viewed as a special case of Lemma~\ref{lem::block_quan} by setting $g=1$. In this case, the vector $\bd$ becomes a scalar $\mathsf{deq}(\mathsf{qtz}([\hat{\bk}_{t-1}]_t)) - [\hat{\bk}_{t-1}]_t$ and the matrix $\bH_{t}$ becomes a vector $\bh_t$. 
\end{proof}

\begin{lem}
\label{lem::rm_block}
Suppose we can write 
\begin{align*}
    \bA_{t+1}^{-1}
    = \begin{bmatrix}
        \bM_{t+1} & \bN_{t+1}^T\\
        \bN_{t+1} & \bO_{t+1}
    \end{bmatrix},
\end{align*}
where $\bM_{t+1} \in \Reals^{tg\times tg}$, $\bN_{t+1} \times \Reals^{g\times tg}$, $\bO_{t+1}\in \Reals^{g\times g}$. Then we have
\begin{align*}
\bA_{t}^{-1}
= \bM_{t+1} - \bN_{t+1}^T \bO_{t+1}^{-1} \bN_{t+1}.
\end{align*}
\end{lem}
\begin{proof}
This lemma can be obtained directly by applying Corollary~\ref{cor::rem}.
\end{proof}

\begin{proof}[Proof of Proposition~\ref{prop::rem_one}]
This proposition is a special case of Lemma~\ref{lem::rm_block} and can alternatively be derived using Lemma~1 from \citet{frantar2022optimal}. 
\end{proof}

\begin{rem}
Assume $T = d/g$ is an integer. By definition, $\bA_{T} = \bP_{\text{inv}} = (\bI + \lambda \hat{\bQ}^T \hat{\bQ})^{-1}$, which implies that $\bA_{T}^{-1} = \bI + \lambda \hat{\bQ}^T \hat{\bQ}$. Then, by using Lemma~\ref{lem::rm_block}, we can iteratively compute $\bA_{T-1}^{-1}, \dots, \bA_{1}^{-1}$.
\end{rem}

\begin{prop}
\label{prop::comp_reduce}
The iterative algorithm based on Proposition~\ref{prop::upd_key} has a computational complexity of $O(d^4)$. By applying Proposition~\ref{prop::rem_one} to update $\bA^{-1}_t$, this complexity is reduced to $O(d^3)$.
\end{prop}
\begin{proof}
The matrix $\bI + \lambda \hat{\bQ}^T \hat{\bQ}$ has dimensions $d \times d$, and computing its inverse to obtain $\bP_{\text{inv}}$ has a complexity\footnote{If $r$ is significantly smaller than $d$, the Woodbury identity can be applied to reduce this complexity to $O(r d^2)$.} of $O(d^3)$. 
For $\bA_t \in\Reals^{t \times t}$, directly computing its inverse requires $O(t^3)$ operations. Additionally, computing $\bB_t \bh_t$ involves $O((d - t)t)$ operations. Since the iterative algorithm processes $t = 1, \dots, d$, the total computational cost is:
\begin{align*} 
O\left(d^3 + \sum_{t=1}^d t^3 + (d - t)t \right) = O(d^4). \end{align*}
By leveraging Proposition~\ref{prop::rem_one}, the cost of updating $\bA_t^{-1}$ reduces to $O(t^2)$ per iteration, decreasing the overall computational complexity to $O(d^3)$.
\end{proof}

\section{Memory Requirements for KV Cache Management}
\label{append::kvmemory}

Maintaining a KV cache requires substantial memory. Specifically, for a batch of $b$ queries, each with a sequence length of $(l_{\text{prompt}} + l_{\text{response}})$, an LLM with $L$ layers, $h$ attention heads per layer, hidden dimension $d$, and precision $p$ (e.g., 2 bytes per parameter for FP16) requires $\text{memory} = 2 \times b \times (l_{\text{prompt}} + l_{\text{response}}) \times L \times h \times d \times p \text{ bytes}$ for storing the KV cache. 
In the case of an LLaMA-2 7B model with a batch of 4 queries and a context length of 2048, the memory requirement is $2 \times 4 \times 2048 \times 32 \times 32 \times 128 \times 2 \text{ bytes} \approx 4 \text{ GB}$.

\section{Details of Our Main Algorithm}
\label{append::algorithm}

\begin{figure*}[t]
    \centering
    \includegraphics[width=0.81\linewidth]{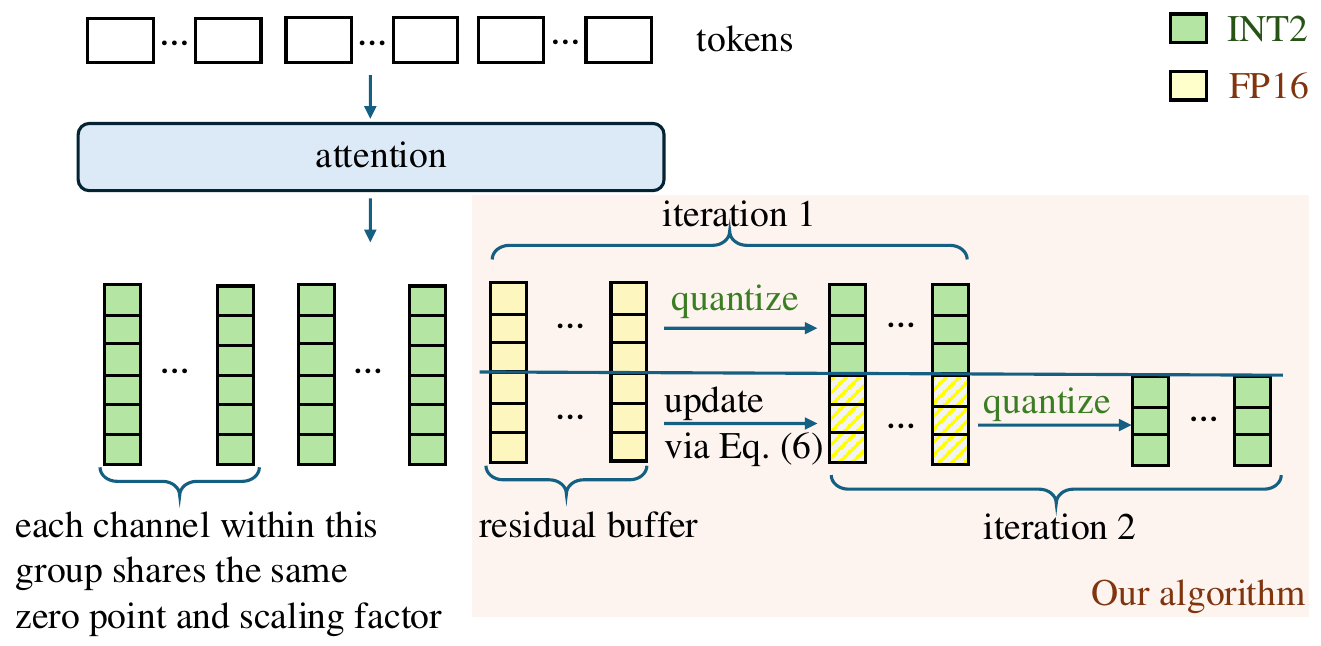}
    \caption{We illustrate how our iterative algorithm quantizes key tensors. We maintain a residual buffer to store the most recent tokens' key tensors in their original precision. Once the buffer reaches a predefined size (e.g., 32 in our experiments), we apply our quantization algorithm to all stored tensors. Specifically, we first quantize a block of elements (e.g., the first half of the hidden dimensions), then update the remaining elements using Eq.~\eqref{eq::update_block}. This process continues iteratively until all elements are quantized.
    }
    \label{fig:our_algo}
\end{figure*}

We present the pseudocode of our main algorithm in Algorithm~\ref{alg::main_alg}. As our main focus is on quantizing key tensors, we describe only this process. For value tensors, we adopt a similar procedure with Algorithm~1 in \citet{liu2024kivi}. Specifically, we maintain a residual buffer that stores the most recent tokens' value tensors in their original data type (e.g., FP16). When the residual buffer reaches a size of $R+1$, we quantize the oldest token's value tensor and remove it from the buffer. Value tensors are quantized per token, with each token's hidden dimension $d$ divided into $\lceil d/G \rceil$ groups of size $G$. For the sake of illustration, we assume the hidden dimension $d$ is divisible by the number of elements quantized per iteration $g$ the group size $G$; and that the residual buffer length $R$ is divisible by $G$.

Figure~\ref{fig:our_algo} illustrates how our algorithm operates. Once the residual buffer reaches a predefined size, we apply our quantization algorithm to all stored tensors. We begin by quantizing a block of elements, then update the remaining elements using Eq.\eqref{eq::update_block}, repeating this process until all elements are quantized. Compared to Eq.\eqref{eq::opt_update}, Eq.~\eqref{eq::update_block} is a more general formulation that supports block-wise quantization rather than quantizing one element at a time.
For illustration, the figure assumes that the residual length equals the group size, though our method allows for setting the group size smaller than the residual length to enable finer-grained quantization.

In the pseudocode, we present our algorithm using multi-head attention. However, the method naturally extends to grouped-query attention. For example, our experiments with \texttt{Llama-3.1-8B-Instruct}---which uses grouped-query attention with 8 KV heads and 32 attention heads---follow the same procedure. The only modification is in Line~5: for each KV head, we concatenate the query tensors from its associated query heads before applying SVD.

We assume a batch size of $1$ in the pseudocode. When the batch size exceeds $1$, we share the same matrices $\bA_{1}^{-1},\dots,\bA_{T-1}^{-1}, \bP_{\text{inv}}$ across all samples in the batch, computing them from the first sample. This significantly reduces memory usage for large batch sizes. Empirically, we find that this does not degrade LLM performance compared to computing separate matrices for each sample. This aligns with our observation in Section~\ref{subsec::query_space} that query tensors tend to lie within a small subspace, even when constructed from different sequences in the same dataset.

\begin{algorithm}[httb]
\begingroup
\small
\caption{Main Algorithm for Key Tensor Quantization}
\label{alg::main_alg}

\begin{algorithmic}[1]

\State \textbf{Input:} Residual length $R$, group size $G$, number of elements to quantize per iteration $g$, hidden state dimension $d$, hyper-parameter $\lambda$, subspace dimension $r$

\Statex

\State \textit{// Prefill Phase}

\State \textbf{Input:} $\bX \in \Reals^{l_{\text{promp}} \times d}$

\State $\bQ = \bX \bW^{Q}$, $\bK = \bX \bW^{K}$

\State $\_, \bSigma, \bV = \mathsf{SVD}(\bQ)$ \Comment{can be accelerated using block power method \citep{bentbib2015block}}

\State $\hat{\bQ} = \diag(\bSigma[:r]) \bV[:r]$

\State $T = d / g$ \Comment{total iterations required to run our algorithm}

\State Compute $\bA_{T}^{-1} = \bI + \lambda \hat{\bQ}^T \hat{\bQ}$ and $\bP_{\text{inv}} = (\bI + \lambda \hat{\bQ}^T \hat{\bQ})^{-1}$

\State Apply Lemma~\ref{lem::rm_block} to obtain $\bA_{T-1}^{-1},\cdots, \bA_1^{-1}$

\State $r = l_{\text{promp}} \% R$,~~$\text{num\_group} = (l_{\text{promp}} - r) / G$

\State Initialize $\bK^{\text{quant}} = []$ and set $\bK^{\text{full}} = \bK[l_{\text{promp}}-r:]$

\State \textit{// Quantize each group of keys (run in parallel)}
\For{$i = 1$ to $\text{num\_group}$}
    \State $\bK_g = \bK[(i-1)G + 1 : iG]$ \Comment{elements in the $i$-th group to be quantized}
    \State $\bK_g^\text{quant} \gets \Call{Quant}{\bK_g, \bA_{1}^{-1}, \cdots, \bA_{T-1}^{-1}, \bP_{\text{inv}}}$
    \State Append $\bK_g^\text{quant}$ to $\bK^{\text{quant}}$
\EndFor

\Statex

\State \textit{// Decoding Phase}

\State \textbf{Input:} $\bx \in \Reals^{1 \times d}$

\State $\bk = \bx \bW^{K}$ and Append $\bk$ to $\bK^{\text{full}}$

\State $\text{num\_group} = R / G$

\If{$\text{len}(\bK^{\text{full}}) = R$}
    
    \For{$i = 1$ to $\text{num\_group}$}
        \State $\bK_g = \bK^{\text{full}}[(i-1)G + 1 : iG]$
        \State $\bK_g^\text{quant} \gets \Call{Quant}{\bK_g, \bA_{1}^{-1}, \cdots, \bA_{T-1}^{-1}, \bP_{\text{inv}}}$
        \State Append $\bK_g^\text{quant}$ to $\bK^{\text{quant}}$
    \EndFor
    \State Reset $\bK^{\text{full}} = []$
\EndIf

\Statex

\Function{Quant}{$\bK$, $\bA_{1}^{-1},\cdots,\bA_{T-1}^{-1}$, $\bP_{\text{inv}}$}
    \For{$t = 1$ to $T$}
        \State Quantize $\bK[:,(t-1)g+1: tg]$ per channel
        \If{$t < T$}
            \State Update $\bK[:,tg:]$ using Lemma~\ref{lem::block_quan} with $\bA_{t}^{-1}$ and $\bP_{\text{inv}}$
        \EndIf
    \EndFor
    \State \textbf{Return:} Quantized $\bK$
\EndFunction

\end{algorithmic}
\endgroup
\end{algorithm}

\section{More Experimental Results}

\subsection{Experimental Setup for Figure~\ref{fig:subspace}}
\label{append::semi_syn}

\begin{figure*}[t]
    \centering
    \includegraphics[width=0.30\linewidth]{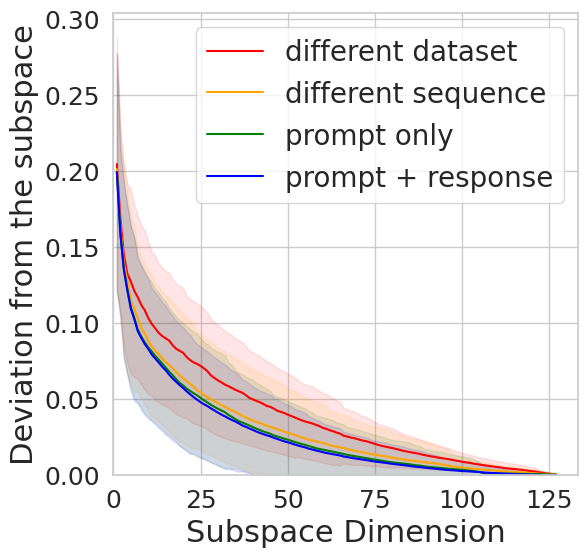}
    \hspace{0.2cm}
    \includegraphics[width=0.30\linewidth]{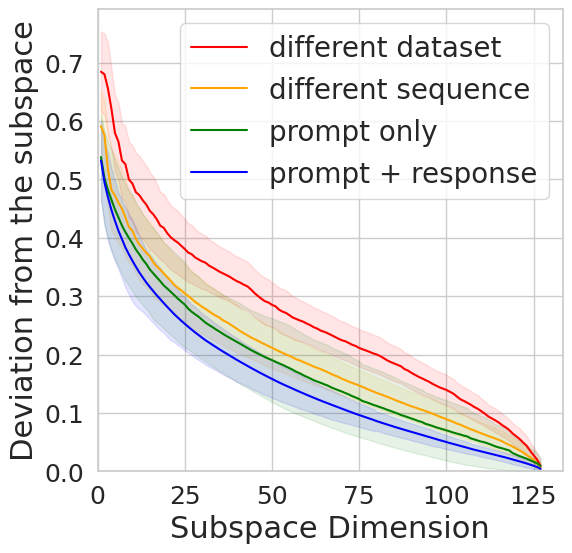}
    \includegraphics[width=0.30\linewidth]{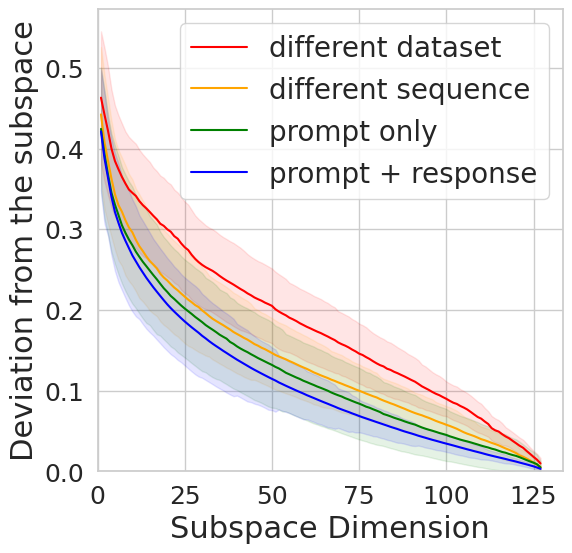}
    \caption{We reproduce the results from Figure~\ref{fig:subspace} (left) for query tensors at different attention layers: the 1st layer (left), the 24th layer (middle), and the 32nd layer (right).
    }
    \label{fig:subspace_different_head}
\end{figure*}

We describe the experimental setup for the results presented in Figure~\ref{fig:subspace}. Unlike the experiments in Section~\ref{sec::exp}, Figure~\ref{fig:subspace} is based on a semi-synthetic setup. Specifically, we first use the \texttt{Llama-3.1-8B-Instruct} model to generate responses \emph{without} KV cache quantization during decoding, ensuring that quantization errors do not alter the response trajectory. Additionally, we analyze query and key tensors before applying rotary position embeddings (RoPE) \citep{su2024roformer} to isolate quantization effects from positional encoding. Figure~\ref{fig:subspace} (left) shows the deviation error for query tensors at the 12th attention layer. To further validate our findings, Figure~\ref{fig:subspace_different_head} extends these results across different layers.

\subsection{Experimental Setup for Section~\ref{sec::exp}}
\label{append::setup}

For SQuat, SQuat\textsuperscript{pre}, we set the group size (i.e., the number of elements quantized together) to 32 and the residual length (i.e., the number of most recent tokens whose KV cache is stored in full precision) to 32. The KV cache is quantized to 2 bits. This configuration provides the best throughput and minimal memory usage (see Figure~\ref{fig:mem_throughput}). For LM-Eval tasks and all LLMs, we set the subspace dimension to $r=5$, the regularization coefficient to $\lambda=0.001$, and the number of elements to quantize per iteration to $g=64$, resulting in just two iterations of our algorithm. For LongBench tasks, which involve longer inputs, we adjust the hyper-parameters as follows. For \texttt{Llama-2-7B-hf}, we use $r=5$, $\lambda=0.001$, $g=32$ for SQuat\textsuperscript{pre}, and $r=40$, $\lambda=0.001$, $g=64$ for SQuat. For \texttt{Llama-3.1-8B-Instruct}, we set SQuat\textsuperscript{pre} to $r=60$, $\lambda=0.0005$, $g=32$, and SQuat to $r=10$, $\lambda=0.0005$, $g=64$. For \texttt{Mistral-7B-Instruct-v0.3}, we use $r=20$, $\lambda=0.0005$, $g=64$ for SQuat\textsuperscript{pre}, and $r=5$, $\lambda=0.001$, $g=64$ for SQuat. For our experiments using \texttt{DeepSeek-R1-Distill-Llama-8B} on the \texttt{math-500} dataset (Figure~\ref{fig:math500}), we set $r=40$, $\lambda=0.0005$, and $g=64$ for both SQuat and SQuat\textsuperscript{pre}.

For KIVI and GEAR, we use the same configuration: a group size of 32, a residual length of 32, and 2-bit quantization for the KV cache.
For KIVI, we use the official implementation from their GitHub \href{https://github.com/jy-yuan/KIVI}{repository}. For GEAR, we set \texttt{rank} = \texttt{rankv} = 2 and \texttt{loop} = 3, following the recommendations in their test script. Since their CUDA-based implementation does not support the Mistral model, we compare only on LLaMA models using their implementation from GitHub \href{https://github.com/opengear-project/GEAR}{repository}. For ZipCache, we quantize important tokens to 4 bits and unimportant tokens to 2 bits, with \texttt{streaming\_gap} = 32 to match the setup used for other baselines and our method. We set the \texttt{unimportant\_ratio} to 0.6 for \texttt{Llama-2-7B} on LM-eval tasks, 0.7 on LongBench tasks, 0.5 for \texttt{Llama-3.1-8B-Instruct} on LM-eval tasks, and 0.6 on LongBench tasks. These values ensure that their KV cache size is roughly aligned with that of other baselines, allowing for a fair comparison. As ZipCache is only implemented for LLaMA in their GitHub \href{https://github.com/ThisisBillhe/ZipCache?tab=readme-ov-file}{repository} (as of the date of our paper submission), we include it in comparisons for LLaMA models only.

For LM-Eval, since \texttt{MMLU\_Pro\_Math} and \texttt{MMLU\_Pro\_Law} all come from \texttt{MMLU\_Pro}, we compute the weighted average in the follow way:
\begin{align*}
\texttt{Average} = \frac{1}{5}\texttt{GSM8k} + \frac{1}{10}(\texttt{MMLU\_Pro\_Math} + \texttt{MMLU\_Pro\_Law}) + \frac{1}{5} \texttt{IFEval} + \frac{1}{5} \texttt{GPQA} + \frac{1}{5} \texttt{BBH}.
\end{align*}
For LongBench, we follow the recommendation in \citet{liu2024kivi} and set the maximum sequence length to 4,096 for \texttt{Llama-2-7B} and 8,192 for all other models. We report the KV size as the average percentage of the compressed cache relative to the FP16 cache at the end of the sequence. It includes all information needed to recover or dequantize the KV cache back to its original data types.
All our experiments are conducted on a single Nvidia H100 GPU (80GB).

\subsection{Ablation Study}
\label{append::ablation}

\begin{table}[t]
\small
\centering
\renewcommand{\arraystretch}{1.25}
\begin{tabular}{lccccc}
\toprule
& $\lambda = 1\times 10^{-4}$ & $\lambda = 5\times 10^{-4}$ & $\lambda = 1\times 10^{-3}$ & $\lambda = 1\times 10^{-2}$ & $\lambda = 1\times 10^{-1}$ \\
\midrule
$r = 5$ & 71.72\% & 73.09\% & 72.71\% & 70.96\% & 4.47\%\\
$r = 10$ & 72.93\% & 71.57\% & 71.65\% & 70.81\% & 6.98\%\\
$r = 20$ & 73.62\% & 72.71\% & 71.27\% & 69.22\% & 5.61\%\\
$r = 40$ & 72.63\% & 72.48\% & 72.18\% & 66.95\% & 6.67\%\\
$r = 60$ & 73.24\% & 72.56\% & 71.87\% & 68.68\% & 16.83\%\\
\bottomrule
\end{tabular}
\caption{Ablation study on the GSM8k dataset using the \texttt{Llama-3.1-8B-Instruct} model. We vary the regularization weight $\lambda$ and the subspace dimension $r$ in SQuat.}
\label{table:lambda_subspace_grid}
\end{table}

We conduct an ablation study on two most important hyper-parameters in our algorithm: the regularization weight $\lambda$ in \eqref{eq::gen_upd_opt_s}, and the subspace dimension $r$ of the query matrix $\hat{\bQ}$. Experiments are performed on the GSM8k dataset using the \texttt{Llama-3.1-8B-Instruct} model.

Recall that the hyper-parameter $\lambda \in [0,\infty)$ controls the trade-off between preserving the original key vectors during (de)quantization and enforcing orthogonality of the residuals to the query subspace. When $\lambda = 0$, the algorithm simplifies to a compression-based quantization method. The query subspace $\hat{\bQ} \in \Reals^{r \times d}$ has top $r$ important query directions. 
Results are summarized in Table~\ref{table:lambda_subspace_grid}. We find that performance degrades when $\lambda > 10^{-3}$. This is because $\hat{\bQ} = \diag(\bSigma[:r]) \bV[:r]$ (Line~6 in Algorithm~\ref{alg::main_alg}) and the leading singular values $\bSigma[:r]$ are typically large. Empirically, we observe that $\lambda = 0.001$ and $r = 5$ consistently yield strong performance. Although this configuration does not achieve the best score on GSM8K, we adopt it as the default for both SQuat and SQuat\textsuperscript{pre} across all models and all tasks in LM-Eval.

\end{document}